\documentclass{article}
\usepackage{spconf,amsmath,graphicx,algorithm,algorithmic,subfigure,multicol,multirow,amssymb,bm}
\usepackage{array}
\usepackage{booktabs}

\title{FACE SPOOFING DETECTION BY FUSING BINOCULAR DEPTH AND SPATIAL PYRAMID CODING MICRO-TEXTURE FEATURES}

\name{Xiao Song, Xu Zhao*\thanks{This research has been supported by the funding from NSFC (61673269,61375019, 61273285). * indicates corresponding author. }, Tianwei Lin}
\address{Key Laboratory of System Control and Information Processing MOE\\
Department of Automation, Shanghai Jiao Tong University}

\begin{document}

\maketitle
\begin{abstract}

Robust features are of vital importance to face spoofing detection, because various situations make feature space extremely complicated to partition. Thus in this paper, two novel and robust features for anti-spoofing are proposed. The first one is a binocular camera based depth feature called Template Face Matched Binocular Depth (TFBD) feature. The second one is a high-level micro-texture based feature called Spatial Pyramid Coding Micro-Texture (SPMT) feature. Novel template face registration algorithm and spatial pyramid coding algorithm are also introduced along with the two novel features. Multi-modal face spoofing detection is implemented based on these two robust features. Experiments are conducted on a widely used dataset and a comprehensive dataset constructed by ourselves. The results reveal that face spoofing detection with the fusion of our proposed features is of strong robustness and time efficiency, meanwhile outperforming other state-of-the-art traditional methods.
\end{abstract}
\section{Introduction}
\label{sec:intro}
Recently, face spoofing detection develops into a deeply concerned research topic in computer vision field. Most of anti-spoofing methods existed can be categorized into three chief varieties: physiological sign based approaches, texture based approaches and illumination peculiarity based approaches. A good survey of research against spoofing attacks can be found in \cite{nixon2008spoof,pan2008liveness}. Physiological sign based methods aim at capturing biometric motions such as eye blinking \cite{sun2007blinking,pan2007eyeblink}, mouth movements \cite{kollreider2007real} and the holistic facial motions \cite{kollreider2009non,bao2009liveness}, etc. However, most of the methods mentioned above are performed in face image sequences only, and may fail when attacked by movable 3D models or videos of live people.

Another generally used facial cues for spoofing detection is illumination peculiarity \cite{kim2009masked,pinto2015face}. For instance, Zhang \emph{et al} \cite{zhang2011face} analyze the different multi-spectral reflectance distributions between genuine and fake faces using Lambertian model. Some researchers merely utilize the illumination cue from a single image \cite{li2004live,tan2010face}. For instance, Tan \emph{et al} \cite{li2004live} employ Retinex-based method to extract illumination reflectance feature for classification. However, these approaches always need extra devices and may fail when attacking photos or prints are of high qualities.

In \cite{yang2013face}, it is  demonstrated that local micro-texture is a useful cue when attacked by recaptured images. Maatta \emph{et al} \cite{maatta2012face} analyze micro-texture features with diverse operators. In \cite{maatta2011face}, the same authors present a novel micro-texture descriptor called Multi-Scale Local Binary Patterns (MSLBP). Freitas \emph{et al} \cite{de2012lbp} propose LBPTOP operator by fusing space information with time information. However, the micro-texture feature is low-level thus they are sensitive to intense illumination change and prints or photos of high qualities.

\begin{figure*}[tb]
\centering
\includegraphics[width=6.05in,height=3.0in]{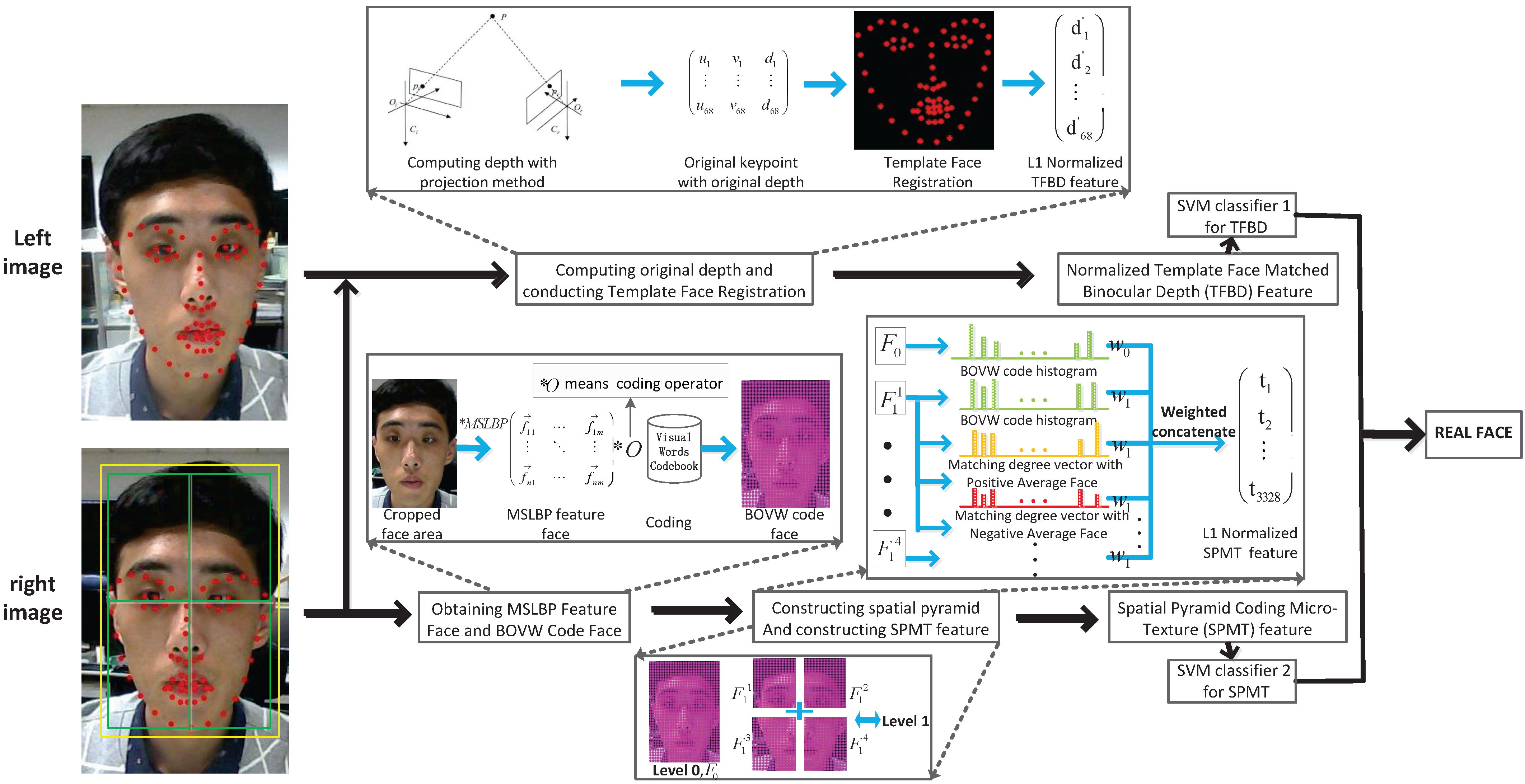}
\caption{The architecture of face spoofing detection with our proposed features}
\end{figure*}

However, most above-mentioned methods are based on low-level descriptors and barely exploit the information derived from rigid structure of live face. The main contribution of our work is that two novel features are proposed for multi-modal classification. Binocular camera system is adopted to calculate original depth value. Every detected facial key point is augmented with the third dimension of depth, and then transformed by our proposed registration transformation to match corresponding key point in template face for several iterations. Afterwards Template Face Matched Binocular Depth feature vector is constructed. MSLBP \cite{maatta2011face} descriptor is applied to each pixel in facial region and Bag of Visual Word (BOVW) code \cite{csurka2004visual} per pixel is obtained using pre-trained codebooks. Then our spatial pyramid coding method is implemented, where weighted and normalized BOVW code histograms, as well as matching-degree vectors matching with two average intra-class faces from corresponding sub-regions, are concatenated as Spatial Pyramid Coding Micro-Texture feature vector. The system architecture is shown in Fig. $1$.

We set up nonparallel dual cameras and conduct stereo calibration, obtaining two images at the same time, from which TFBD feature and SPMT feature are extracted. Each feature vector is individually fed to a nonlinear SVM classifier and score-level fusion of two individual SVM outputs determines classification result. 

\section{SPOOFING DETECTION WITH PROPOSED FEATURES}
\label{feature}

\subsection{Template Face Matched Binocular Depth Feature}

The Constrained Local Models (CLM) \cite{cristinacce2008automatic} is adopted to locate $68$ facial key points and obtain their pixel coordinates in both left and right images.

$E$ is defined as rotation matrix and $V$ represents translation vector. $M_{r}=\{f_{xr},c_{xr},f_{yr},c_{yr}\}$,   $M_{l}$  are defined as intrinsic matrices of right and left cameras respectively. $p_{l}=[u_{l},v_{l},1]^{T}$, $p_{r}=[u_{r},v_{r},1]^{T}$ are homogeneous pixel coordinates of certain keypoint in left image and right image.
$m=M_{l}\left[\begin{array}{lcr}E & V  \\0^{T} & 1 \end{array}\right]$.
According to pinhole camera model, original depth value of certain facial keypoint $d$ is obtained:
{\setlength\abovedisplayskip{3pt}
\setlength\belowdisplayskip{-1pt}
\begin{align}
  d&=\frac{B_{12}b_{2}-B_{22}b_{1}}{\frac{u_{r}-c_{xr}}{f_{xr}}(B_{12}B_{21}-B_{11}B_{22})+(B_{12}B_{23}-B_{22}B_{13})}
\end{align}
}
in which $B_{1j}=m_{1j}-m_{3j}u_{l},B_{2j}=m_{2j}-m_{3j}v_{l},b_{1}=m_{34}u_{l}-m_{14},b_{2}=m_{34}v_{l}-m_{24}$ are intermediate variables.

Three dimensional abstract keypoint is defined for registration operation where the first and second dimensions are pixel coordinate in the right image and the third dimension is normalized depth value which powerfully reflects stereo structure of face. Each face to be detected can be denoted by a set with 68 abstract keypoints: $\{p_{j}|p_{j}=[x_{j},y_{j},d^{'}_{j}]^{T},1\leq j \leq 68\}$. A set of standard abstract keypoints called Template Face $T$ is needed: $T=\{T_{j}|T_{j}=[T^{j}_{x},T^{j}_{y},T^{j}_{d}]^{T},1\leq j \leq 68\}$. Template face is obtained before training. We select $20$ image pairs collected from $5$ different subjects with moderate distance away from cameras.  Cameras are usually placed directly in front of their faces while pictures are shot. 

$x^{j}_{i},y^{j}_{i}$ are defined as $x$ and $y$ values of the $jth$ abstract keypoint in the $ith$ right picture among 20 selected pairs. $d^{j}_{i}$ and $d^{'j}_{i}$ represent original depth and normalized depth: $d^{'j}_{i}=d^{j}_{i}-(\sum_{j}d^{j}_{i})/68$. Template Face is obtained as follows:
{\setlength\abovedisplayskip{5pt}
\setlength\belowdisplayskip{-5pt}
\begin{align}
  T^{j}_{x}&=\sum_{i}x^{j}_{i}/20, \ T^{j}_{y}&=\sum_{i}y^{j}_{i}/20, \ T^{j}_{d}&=\sum_{i}d^{'j}_{i}/20
\end{align}}

Single-round registration paradigm is defined as $\hat{p}_{j}=s\cdot R\times p_{j}+t$, where $s$ represents scaling factor, $R,t$ indicate rotation matrix and translation vector for abstract keypoint. Every round of registration transformation seeks for optimal parameters with minimum registration error:
{\setlength\abovedisplayskip{1pt}
\setlength\belowdisplayskip{1pt}
\begin{align}
  (s^{*},R^{*},t^{*})&=\emph{arg}min\sum_{j=1}^{68}{\parallel T^{j}-s\cdot R\times p_{j}-t\parallel}^{2}
\end{align}}
Absolute orientation using unit quaternions algorithm \cite{horn1987closed} is adopted to solve Eq.$(3)$. Every abstract key point $\hat{p}_{j}$ can be solved with  $(s^{*},R^{*},t^{*})$ after single round of registration.

An optimized iteration method is proposed based on Iterative Closest Point algorithm. We collect $20$ abstract keypoints with minimum errors from current round of registration as parameters, to search optimal parameters $(s^{*},R^{*},t^{*})$ in the next round. After $20$ rounds of registraion, $68$ depth values extracted from $68$ abstract keypoints is concatenated as $68$ dimensional TFBD Feature vector.

\subsection{Spatial Pyramid Coding Micro-Texture Feature}
Anti-spoofing method based on only TFBD feature may fail when three dimensional structures of fake faces are highly similar to real faces. Hence, high-level micro-texture feature is needed to fuse with binocular depth feature.

A cascade detect model \cite{viola2001rapid} is implemented to detect the face region in right image. Before cropped, the detected face area should be expanded because it is demonstrated in \cite{yang2013face} that most discriminative areas locate in marginal areas of face. Afterwards, the cropped face area is normalized into a $64\times72$ pixel gray-scale image.

Basic MSLBP operator using circular neighborhood introduced in \cite{maatta2011face} is adopted. The notation $LBP^{u}_{P,R}$ indicates that $P$ sampled pixels on circle with a radius of $R$ are compared with central pixel of neighborhood and uniform pattern \cite{maatta2011face} is adopted to transform original LBP value, reducing the amount of labels for $LBP_{P,R}$.

A MSLBP operator with a capacity of $3$ illustrated as $\{LBP^{u}_{8,1},LBP^{u}_{8,2},LBP^{u}_{16,2}\}$ is applied per pixel in normalized face image.
MSLBP Feature Face $F^{u}_{mp}$ is defined as low-level texture descriptor. Each ``pixel'' in $F^{u}_{mp}$ has three channels: $F^{u}_{mp}(x,y)=[LBP^{u}_{8,1}(x,y),LBP^{u}_{8,2}(x,y),LBP^{u}_{16,2}(x\\,y)]$. Values of three LBP used are $8$-bit long after transformed with uniform pattern.

A medial-level texture descriptor called BOVW code face $F_{bw}$ is introduced. A MSLBP Feature codebook with a capacity of $256$ is obtained by K-means clustering algorithm before training and testing, where we select $3000$ representative MSLBP feature faces from training set for clustering. Codebook can be notated as: $\{code_{i}|code_{i}=[code^{1}_{i},code^{2}_{i},code^{3}_{i}]\}$ for $1\leq i \leq 256$. Afterwards BOVW coding algorithm in Eq. $(4)$ is applied per ``pixel'' in MSLBP feature Face, obtaining BOVW code face with the size of $64\times72$:
{{\setlength\abovedisplayskip{1pt}
\setlength\belowdisplayskip{-3pt}
\begin{align}
  F_{bw}(x,y)&=\emph{arg}min_{k}\sum_{n=1}^{3}{\parallel code^{n}_{k}-F^{u}_{mp}(x,y)[n]\parallel}^{2}
\end{align}}

Inspired by Spatial Pyramid Matching \cite{lazebnik2006beyond}, a novel spatial pyramid coding algorithm is proposed. BOVW Code Face preserves original spatial layout thus it can be partitioned. In the first step, two-level pyramid is constructed on $F_{bw}$ where level $l$ means each spatial dimension is subdivided by factor $2^{l}$ for $l=0,1$. Let $F^{1}_{0}$ notate whole BOVW code face, $F^{i}_{1}$ for $i=1,2,3,4$ notate the $ith$ sub-region under level $1$. Thus $5$ sub-regions are obtained. The higher level of resolution is, the greater weight sub-region owns. Weight pyramid is adopted with $w_{l}=\frac{1}{2^{L-l}}$, where $w_{l}$ notates partitioned sub-region's weight under level $l$ and $L$ represents maximal level of resolution. Each normalized BOVW code histograms constructed from corresponding sub-region is weighted by $w_{l}$.

In the second step, two datasets independent from training and test set are collected and named as positive and negative average sets, in which $2000$ positive samples and $2000$ negative samples are included  respectively. Then average intra-class face is introduced:
{{\setlength\abovedisplayskip{1pt}
\setlength\belowdisplayskip{1pt}
\begin{align}
  A_{N or P}&=\frac{1}{|\Omega_{N or P}|}\sum_{i\in\Omega_{N or P}}F_{bw}^{i}
\end{align}}
where $\Omega_{N or P}$ represents positive or negative average set. Thus Positive Average Intra-class Face $A_{P}$ and Negative Average Intra-class Face $A_{N}$ are obtained.

Matching-degree vector is introduced in Eq.$(6)$, where $\mathbf{1}(\cdot)$ is indicator function, $\gamma=N$,$P$ indicates matching with positive or negative average face, $l$ is resolution level, $c\in[0,512]$ represents BOVW code value, $f^{i}_{l}(c)=\sum_{(x,y)}\mathbf{1}(F^{i}_{l}(x,y)=c)$, $a^{i}_{\gamma,l}(c)=\sum_{(x,y)}\mathbf{1}(A^{i}_{\gamma,l}(x,y)=c)$, $i$ means $ith$ sub-region, so each $M$ is $512$ dimensional.
{{\setlength\abovedisplayskip{3pt}
\setlength\belowdisplayskip{-3pt}
\begin{equation}
M^{i}_{\gamma,l}(c)=\left\{
             \begin{array}{lr}
             w_{l}\times1 \qquad   f^{i}_{l}(c)=0\&\&a^{i}_{\gamma,l}(c)=0 &  \\
             w_{l}\times min(\frac{f^{i}_{l}(c)}{a^{i}_{\gamma,l(c)}},\frac{a^{i}_{\gamma,l}(c)}{f^{i}_{l}(c)}) \quad others
             \end{array}
\right.
\end{equation}}

 Two-level pyramid is constructed on $A_{P}$ and $A_{N}$, obtaining $A_{P,1}^{i}$ and $A_{N,1}^{i}$ under level $1$ for $i=1,2,3,4$. For each sub-region $F^{i}_{1}$, two matching-degree vectors $M_{P,1}^{i}$ and $M_{N,1}^{i}$ matching with $A_{P,1}^{i}$ and $A_{N,1}^{i}$ respectively are computed and L-1 normalized.

Finally, $5$ normalized and weighted BOVW code histograms as well as $8$ normalized and weighted matching-degree vectors are concatenated as $3328$ dimensional Spatial Pyramid Coding Micro-Texture feature vector.

\subsection{Classification}
Once 68 dimensional TFBD feature and 3328 dimensional SPMT feature are obtained,
each feature vector is individually fed to a well trained nonlinear SVM classifier. Score-level fusion of two individual SVM outputs is adopted because effects of TFBD and SPMT features on spoofing detection are independent and of the same order.

\section{Experiments}
\label{experiment}
\subsection{Datasets and Setup}
Publicly available NUAA Photograph Imposter Database \cite{tan2010face} and CASIA-FASD Database \cite{zhang2012face} are employed. Considering no available public binocular camera based dataset for face spoofing detection, we construct our own dataset which is composed of $6000$ image pairs captured with two web cameras. Each pair includes a picture taken by left camera and a picture taken by right camera simultaneously. Our machine is a modern PC with $8$GB RAM and GTX$960$ graphics card.

Dual cameras need to be calibrated before computing depth value thus our dataset is sampled with two fixed and calibrated cameras with resolution  $640\times480$, meanwhile our proposed algorithm is resolution tolerant because of its keypoint and ROI based property.
$15$ different subjects are involved and $200$ image pairs are sampled from each person under different illumination condition. People are also required to raise head, lower head, spin face, sit with different position and varied distance away from cameras.
Meanwhile $20$ images of different faces with high definition are printed on A4 papers, other $5$ faces are printed on photographic papers and another $5$ faces are displayed on an ipad screen. $30$ fake faces are obtained in total and $100$ picture pairs are sampled from each fake face under different illumination condition. For every fake face, we move it horizontally, vertically, back and front and rotate it in depth. Especially for those printed on papers or photos, we also bend them inward and outward.

Two individual SVM classifiers for TFBD feature and SPMT feature need to be trained respectively. $3200$ image pairs ($1600$ positive pairs, $1600$ negative pairs) are used for training of the TFBD SVM classifier, meanwhile other $2800$ image pairs in our dataset form the test set for evaluating our two proposed features. Multiple images from three datasets mentioned above are used to train SPMT SVM classifier. Test set in NUAA database is used to solely test our proposed SPMT feature and make comparisons with other descriptors.

\begin{table}[!hbp]
\small
\vspace*{-15pt}
\centering
\caption{Comparison of texture descriptors on NUAA dataset}
\begin{tabular*}{1\linewidth}{p{1cm}<{\centering}p{1cm}<{\centering}p{1cm}<{\centering}p{1cm}<{\centering}p{1cm}<{\centering}p{1.04cm}<{\centering}}
\toprule
Operator & LPQ & Tan's & Mslbp & Yang's & \textbf{SPMT} \\
 \quad &\cite{ojansivu2008blur} &\cite{tan2010face} &\cite{maatta2011face} &\cite{yang2013face}  &\textbf{(ours)} \\
\midrule
Accuracy & 0.870 & 0.881 & 0.928 & 0.975 & \textbf{0.980}\\
AUC &0.931 &0.941 &0.977 &0.992 &\textbf{0.995}\\
EER &14.8\% &13.9\% &8.0\% &2.2\% &\textbf{2.0\%} \\
\bottomrule
\end{tabular*}
\end{table}

\begin{table}[!hbp]
\small
\vspace*{-25pt}
\centering
\caption{Comparison of texture descriptors on our dataset}
\begin{tabular*}{1\linewidth}{p{1cm}<{\centering}p{1cm}<{\centering}p{1cm}<{\centering}p{1cm}<{\centering}p{1cm}<{\centering}p{1.04cm}<{\centering}}
\toprule
Operator & LPQ & Tan's & Mslbp & Yang's & \textbf{SPMT} \\
 \quad &\cite{ojansivu2008blur} &\cite{tan2010face} &\cite{maatta2011face} &\cite{yang2013face}  &\textbf{(ours)} \\
\midrule
Accuracy & 0.842 & 0.858 & 0.894 & 0.947 & \textbf{0.949}\\
AUC &0.851 &0.871 &0.917 &0.975 &\textbf{0.988}\\
EER &17.0\% &14.2\% &12.9\% &6.1\% &\textbf{5.7\%} \\
\bottomrule
\end{tabular*}
\end{table}

\begin{table}[!hbp]
\small
\vspace*{-25pt}
\centering
\caption{Evaluation of our proposed features on our dataset}
\begin{tabular*}{1\linewidth}{p{1.56cm}<{\centering}p{1.22cm}<{\centering}p{1.22cm}<{\centering}p{1.22cm}<{\centering}p{1.25cm}<{\centering}}
\toprule
Operator & Original & TFBD & SPMT & \textbf{SPMT+}  \\
 \quad &depth &feature &feature &\textbf{TFBD}   \\
\midrule
Accuracy & 0.853 & 0.927 & 0.949 & \textbf{0.990} \\
AUC &0.865 &0.955 &0.988 &\textbf{0.998} \\
EER &14.3\% &8.2\% &5.7\% &\textbf{1.0\%}  \\
\bottomrule
\end{tabular*}
\vspace{-5mm}
\end{table}

\subsection{Experimental Results}
 The first experiment solely evaluates our proposed SPMT feature. Area Under Curve (AUC), Equal Error Rate (EER) and accuracy are adopted as assessment criteria. Comparisons are made with four powerful texture descriptors. For instance, MSLBP \cite{maatta2011face} is the state-of-the-art low level descriptor and Yang's component dependent descriptor \cite{yang2013face} is a good medial-level descriptor. Optimal SVM parameters are used for each descriptor. Experiment is conducted on NUAA database and results are shown in Table $1$. As can be seen, our SPMT descriptor outperforms slightly than Yang's, but not obvious. But in terms of time efficiency, component dependent descriptor runs at $4$ fps while our SPMT descriptor runs at $10$ fps which is $2.5$ times faster.

\begin{figure}
\vspace*{-7pt}
\setlength{\abovedisplayskip}{3pt}
\centering
\includegraphics[width=2.9in,height=2.17in]{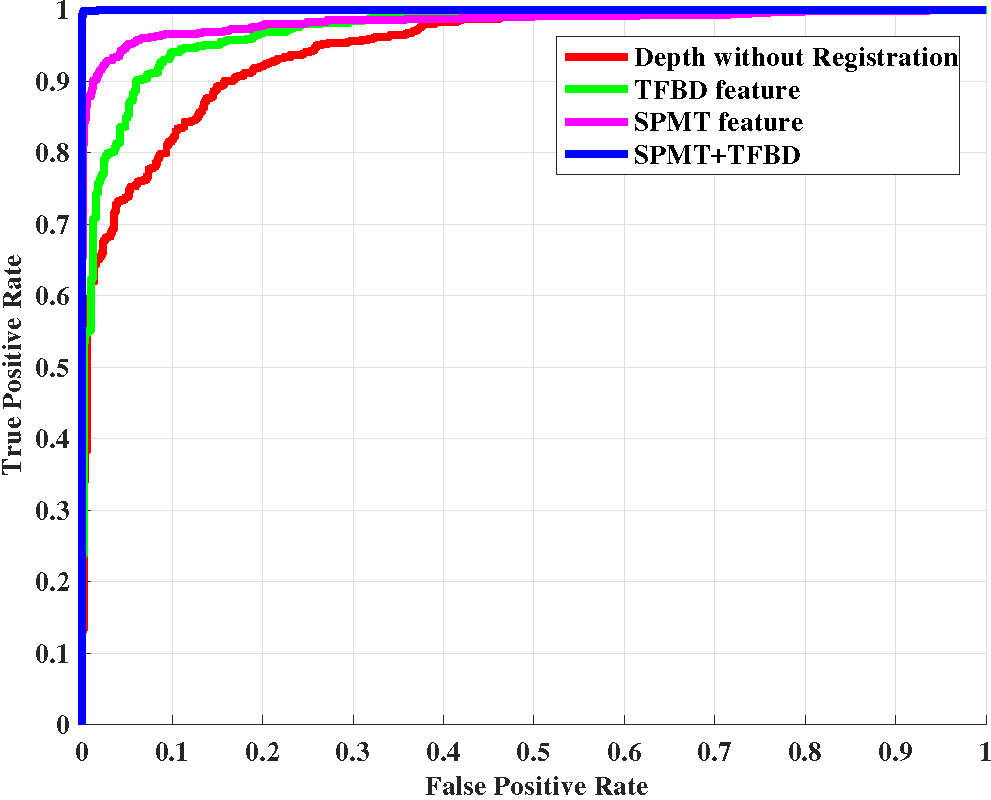}
\vspace{-4mm}
\caption{ROC curves of our proposed feature descriptors evaluated on our dataset}
\vspace{-4mm}
\end{figure}

In NUAA database, cameras always be in front of faces and the distance of the face away from cameras is always moderate. The second experiment is conducted on our own challenging dataset. The results are shown in Table $2$. We can see that all descriptors perform worse but SPMT feature still performs best. For instance, component dependent descriptor needs to extract feature histograms from $4$ components such as left eye, but when face is spun or far away from camera it is hard to locate components precisely. SPMT descriptor aims at excavating global intra-class similarity and spatial distribution of texture, thus sensitivity of feature is reduced.

The third experiment is also conducted on our own dataset. Our TFBD descriptor runs at $50$ fps hence decision level fusion of two features are computational efficient. The results shown in Table $3$ reveal that original depth feature is somewhat discriminative, however when face is increasingly far from cameras, relative differences of depth between different key points are shrinking. Performance also worsens when face in front of camera is spun because of the error caused by imprecise location of key points. After matched with template face, normalized binocular depth feature fully reflects stereo structure of face. $5\%$ improvement in accuracy proves the effectiveness of TFBD feature. But TFBD feature has limitation in its sensitivity to stereo structure which is highly similar to real face. Hence SPMT feature is introduced and its effectiveness is proved by accuracy nearly $95\%$. However, limitation also exists in its sensitivity to intense illumination change and high definition print. Hence, the method of fusing TFBD feature with SPMT feature is adopted eventually. As can be seen, the accuracy is finally improved to $99.2\%$ and AUC value is improved to $0.998$. ROC curves of our proposed features evaluated on our own dataset are shown in Fig. $2$.

\section{Conclusion}
\label{conclusion}


 Proposed TFBD feature highlights differences in three dimensional structures between genuine and fake faces, while the proposed SPMT feature highlights global intra-class similarities and spatial distribution of micro-texture. The multi-modal detection based on these robust features shows comparative advantages on identifying fake faces displayed on photos, prints or videos with high definition. We believe our features can also be applied to other face recognition tasks.


\bibliographystyle{IEEEbib}
\bibliography{strings,refs}

\end{document}